\documentclass{article}

\usepackage{PRIMEarxiv}
\usepackage[utf8]{inputenc}
\usepackage[T1]{fontenc}
\usepackage[hyphens]{url}
\usepackage{graphicx}
\graphicspath{{Figures/}}
\usepackage[numbers,sort&compress]{natbib}
\usepackage{caption}

\usepackage{booktabs}
\usepackage{amsmath}
\usepackage{amssymb}
\usepackage{amsthm}
\usepackage{multirow}
\usepackage{array}
\usepackage{enumitem}
\usepackage{colortbl}
\usepackage{microtype}
\usepackage[colorlinks=true,allcolors=blue]{hyperref}

\definecolor{DwtDctShade}{RGB}{232,242,252}
\definecolor{DwtDctSvdShade}{RGB}{253,241,224}
\definecolor{RivaGANShade}{RGB}{232,247,236}

\DeclareMathOperator{\Embed}{Embed}
\DeclareMathOperator{\Decode}{Decode}
\DeclareMathOperator{\BER}{BER}
\DeclareMathOperator{\Sim}{Sim}
\newcommand{\HFR}{\mathrm{HFR}}
\newtheorem{proposition}{Proposition}
\newtheorem{definition}{Definition}

\title{One Prompt Is Enough: Watermark Laundering Through Foundation Image Models}

\author{
\parbox{0.96\textwidth}{
\centering
Jidong Yang\textsuperscript{1}, Qi Li\textsuperscript{1}, Wei Zong\textsuperscript{2},
Yang-Wai Chow\textsuperscript{2}, Willy Susilo\textsuperscript{2},\\
Huaike Yu\textsuperscript{1}, Chunpeng Wang\textsuperscript{1}, and Suo Gao\textsuperscript{3}\\[0.6em]
{\normalfont\small
\textsuperscript{1}Key Laboratory of Computing Power Network and Information Security, Ministry of Education; Shandong Computer Science Center; Shandong Provincial Key Laboratory of Industrial Network and Information System Security; Shandong Fundamental Research Center for Computer Science; Qilu University of Technology (Shandong Academy of Sciences), Jinan 250353, China\\[0.25em]
\textsuperscript{2}Institute of Cybersecurity and Cryptology (iC2), University of Wollongong, Australia\\[0.25em]
\textsuperscript{3}School of Information Science and Engineering, Dalian Polytechnic University, Dalian 116034, China}\\[0.6em]
{\normalfont\footnotesize\ttfamily
qluliqi@163.com; jidong\_yang\_paper@163.com; mpeng1122@163.com; huaikeyu@gmail.com\\
wzong@uow.edu.au; caseyc@uow.edu.au; wsusilo@uow.edu.au; gaosuodlpu@163.com}
}
}

\begin{document}
\maketitle

\begin{abstract}
Invisible watermarks are typically evaluated against predefined perturbations, such as compression, blur, noise, cropping, and denoising. However, public foundation image models  expose a distinct threat: an attacker can provide a watermarked image with a single reconstruction prompt and obtain a visually faithful output from which the invisible watermark can no longer be decoded reliably. We formalize this failure mode as watermark laundering and evaluate it by evaluate it through a joint payload-fidelity profile that measures bit error rate(BER) together with visual and semantic preservation. Across six OpenAI and Google image editing models, three representative watermarking schemes, and 1,800 reconstructed outputs, we observe two complementary laundering regimes: OpenAI models achieve the strongest attack ability across the evaluated schemes, whereas Nano Banana 2 reveals that DwtDct watermarks remain vulnerable even under high-fidelity reconstruction. Prompt ablations show that no single removal-oriented directive is necessary for attacking watermarks, indicating that the effect is primarily induced by the reconstruction pathway rather than explicit attack prompt. Comparisons with conventional attacks further indicate that prompt-conditioned reconstruction constitutes a distinct operational attack interface, motivating foundation-model reconstruction as a missing robustness condition for invisible watermark evaluation.
\end{abstract}

\keywords{Invisible image watermarking \and Watermark laundering \and Foundation image models \and Provenance security \and Black-box attacks}

\begin{figure}[t]
\centering
\includegraphics[width=.77\linewidth]{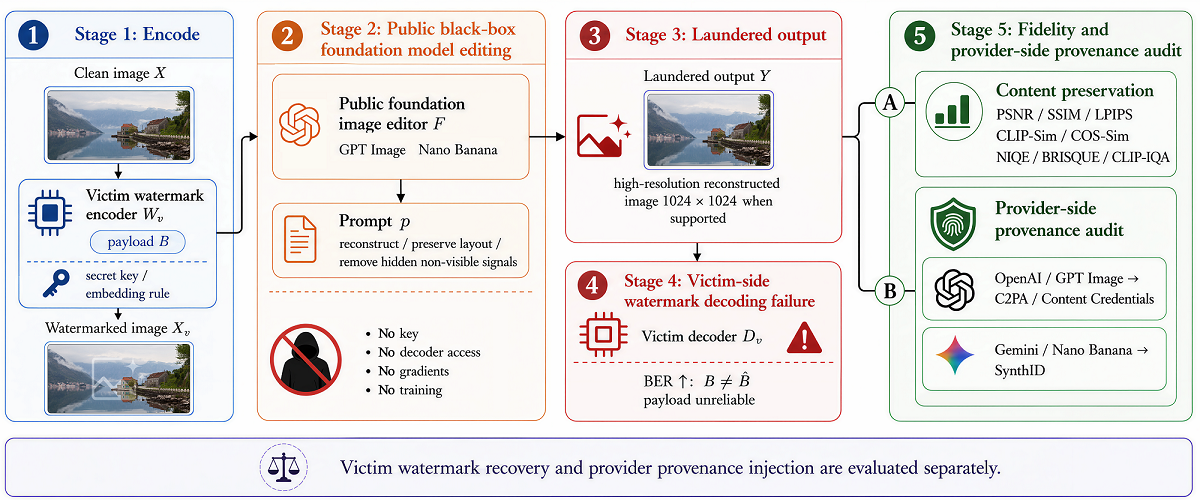}
\caption{Threat model and evaluation for single-prompt watermark laundering through a public black-box foundation image editor. An attacker submits a watermarked image and a single reconstruction prompt without access to the victim decoder, embedding key, model parameters, gradients, or detector feedback. The reconstructed output is evaluated separately for victim-payload recoverability and content preservation.}
\label{fig:watermark-laundering-overview}
\end{figure}

\section{Introduction}
\label{sec:intro}

Invisible watermarks have become a crucial technology for image provenance, copyright protection, and AI-generated content identification. Invisible watermark robustness is usually evaluated using a fixed suite of compression, filtering, geometric transformation, or denoising attacks~\citep{cox1997spread,hartung1999watermarking,cox2007digital}. Recent studies have further extended this evaluation to learning-based removal and regeneration attacks. Such evaluations largely assume the attacker applies an explicit image-processing operator or a dedicated removal model. This assumption becomes increasingly incomplete as image editing capabilities are integrated into foundation image models. An attacker can request a content-preserving reconstruction of watermarked images through a request framed as noise or artifact removal. The output may remain visually and semantically faithful while making the embedded payload unreliable to the victim decoder.

We define this failure mode as \emph{watermark laundering} (Figure~\ref{fig:watermark-laundering-overview}), by analogy with information laundering, in which an intermediary preserves the usable content while weakening its association with the original input~\citep{klein2012informationlaundering,lazer2018sciencefake,vosoughi2018spread,shao2018lowcredibility,starbird2017alternative}. Here, the intermediary is a public foundation image model, and the weakened information is the victim payload. Given a watermarked image, watermark laundering occurs when prompt-conditioned reconstruction preserves its visible or semantic content while making the original watermark unreliable to the victim decoder. Although the reconstructed output may carry provenance metadata conforming to the Coalition for Content Provenance and Authenticity (C2PA) specification, including associated Content Credentials~\citep{c2pa_spec}, such signals neither recover the victim payload nor constitute provenance substitution. Note that payload disruption alone is insufficient: decoder failure caused by severe image destruction is not meaningful laundering, which requires the output to remain visually or semantically usable.

Representative post-hoc watermarking schemes include HiDDeN, StegaStamp, RivaGAN, and MBRS, together with attacks based on signal processing, denoising, learned removal, and generative regeneration~\citep{zhu2018hidden,tancik2020stegastamp,zhang2019rivagan,jia2021mbrs,dabov2007bm3d,zhao2024provably,liu2024ctrlregen}. Unlike prior regeneration attacks that typically require access to a local or controllable generative pipeline, we study deployed public foundation image models invoked through a single natural-language prompt. Therefore, our contribution lies in characterizing a distinct operational threat rather than proposing a new internal removal method. Instruction-guided editing methods further demonstrate that image reconstruction can preserve scene content while rewriting the underlying pixel-level information~\citep{meng2022sdedit,hertz2023prompt2prompt,brooks2023instructpix2pix}.

We systematically evaluate this threat across six public foundation image models from two provider lineages, three representative watermarking schemes, and 1,800 reconstructed outputs. OpenAI models produce the strongest overall payload disruption across the evaluated schemes, while Nano Banana~2 demonstrates that DwtDct remains vulnerable even under high-fidelity reconstruction. Prompt ablations further show that explicit hidden-information-removal wording is not necessary for payload disruption. Requests framed as reconstruction, restoration, or artifact removal may therefore invoke generative processes that weaken invisible evidence, revealing a mismatch between natural-language task framing and low-level forensic preservation. Our experiments study this watermark-laundering effect rather than a bypass of any specific provider policy, and they do not imply any change in legal copyright ownership. Our contributions are summarized as follows:
\begin{itemize}[leftmargin=1.4em,itemsep=1pt,topsep=2pt]
\item We establish a payload--fidelity evaluation framework for watermark laundering under a single-prompt black-box threat model. Rather than introducing a dedicated SOTA attack, we identify foundation-model reconstruction as a missing robustness condition in invisible watermark evaluation.
\item We evaluate six publicly accessible foundation image models and three invisible watermarking schemes across 1,800 outputs generated from a stratified set of 100 images, revealing complementary high-disruption and high-fidelity laundering regimes.
\item We characterize watermark laundering through prompt ablations, comparisons with conventional attacks, and high-frequency residual (HFR) analysis, showing that payload disruption is largely insensitive to the tested prompt components and only partially explained by high-frequency changes.
\end{itemize}

\section{Threat Model and Theoretical Framework}
\label{sec:formalization}

Let $W_V=(\Embed_{W_V},\Decode_{W_V})$ be a victim watermarking scheme. Given a clean image $x$ and a $k$-bits payload $b\in\{0,1\}^k$, the corresponding watermarked image is $x_V=\Embed_{W_V}(x,b)$. A black-box foundation model editor $F$ takes a natural-language prompt $p$ as input and returns the reconstructed output $y=F(x_V,p)$.

\begin{definition}[Watermark laundering]
\label{def:watermark-laundering}
The transformation $x_V\mapsto y$ is watermark laundering when $\hat b=\Decode_{W_V}(y)$ is an unreliable estimate of $b$ while $y$ remains visually faithful to $x_V$ and preserves its semantic content. For an evaluation set $D=\{(x_i,b_i)\}_{i=1}^{N}$, let $x_{V,i}=\Embed_{W_V}(x_i,b_i)$ and $y_i=F(x_{V,i},p)$. We report the following joint profile:
\begin{equation}
\begin{aligned}
\mathcal{P}_D(W_V,F,p)=\bigl(&\overline{\BER}_D,
\overline{M}_{{\rm ref},D},\overline{M}_{{\rm clean},D},\\
&\overline{M}_{{\rm sem},D},\overline{M}_{{\rm nr},D}\bigr).
\end{aligned}
\label{eq:profile}
\end{equation}
where each overbar denotes an average over $D$; $M_{\rm ref}$ measures fidelity to the watermarked input, $M_{\rm clean}$ measures fidelity to the clean image, $M_{\rm sem}$ measures semantic preservation, and $M_{\rm nr}$ measures no-reference image quality. We report raw BER. Random binary recovery corresponds to BER $=0.5$, so movement toward $0.5$ from either direction indicates stronger disruption. A value above $0.5$ is not treated as stronger than random recovery. Fidelity metrics are reported separately because their directions cannot be represented correctly by a single aggregate score.
\end{definition}

Our formulation imposes three conditions. First, the adversary acts through a natural-language instruction rather than training, gradients, or optimization informed by the decoder. Second, the transformation is reconstructive: the editor returns newly rendered pixels rather than a tuned additive perturbation. Third, laundering has two requirements. Payload disruption without content preservation is ordinary destructive removal, whereas preservation without payload disruption is a benign edit. Define the bounded dataset-level removal score $R_{\rm BER}(D)=1-2|\overline{\BER}_D-0.5|$, for which $R_{\rm BER}(D)=1$ denotes random bit recovery. For a prespecified application-specific similarity function $\Sim$, oriented so that larger values indicate stronger content preservation, define $\overline{\Sim}_D=|D|^{-1}\sum_{i=1}^{N}\Sim(y_i,x_{V,i})$. A conceptual binary criterion can then be written as
\begin{equation}
R_{\rm BER}(D)\geq\tau_{\rm RR}
\quad\text{and}\quad
\overline{\Sim}_D\geq\tau_{\rm sim}.
\label{eq:tau-laundering}
\end{equation}
where $\tau_{\rm RR}$ and $\tau_{\rm sim}$ are application-specific thresholds. The abstract function $\Sim$ is a conceptual criterion rather than an aggregate of the reported fidelity metrics. We do not instantiate universal values for these thresholds or report a thresholded laundering success rate. The experiments instead use the continuous joint profile because a single cutoff would obscure the BER--fidelity tradeoff.

The attacker has ordinary black-box API access to $F$ but no knowledge of the victim algorithm, payload, or key. The attacker receives no feedback from the victim decoder or detector and has no access to model training or gradients. The attacker submits a watermarked image and a single prompt to a public editing interface, which returns a reconstructed image rather than a tuned local perturbation. Success requires the returned image to remain usable and related to the source while making recovery of the victim payload unreliable.

Our primary outcome is victim watermark laundering. Let $L$ denote the conceptual conjunction in Equation~\eqref{eq:tau-laundering} under application-specific thresholds, and let $E_{\rm prov}\in\{0,1\}$ indicate whether the returned output contains verified provider provenance evidence. We reserve \emph{provenance substitution} for the stronger state
\begin{equation}
S=L\wedge E_{\rm prov}.
\label{eq:provenance-substitution}
\end{equation}
Thus, provider metadata or a provider watermark cannot substitute for decoding evidence from the victim watermark. Conversely, $L$ alone does not show that the image has acquired a new machine-readable origin. If verified provider evidence is present, an output satisfying state $S$ may present an \emph{apparent} replacement of source or ownership evidence at the interface level, but it does not transfer legal copyright.

We use a behavioral bottleneck abstraction rather than claiming knowledge of proprietary implementations. Let $Z=E_F(x_V)$ denote the representation used for reconstruction, and let $y=R_{F,p}(Z,\xi)$ denote the rendered output. Let $C=C(x)$ summarize objects, layout, scene structure, and visibly relevant appearance. Here, $I(U;V)$ denotes mutual information, and $I(U;V\mid T)$ denotes conditional mutual information, both measured in bits. The account makes three assumptions. First, semantic sufficiency requires $Z$ to retain enough information about $C$ to reconstruct the scene. Second, forensic insufficiency requires $I(b;Z\mid C)$ to be small, meaning that after conditioning on the visible content $C$, the representation $Z$ retains little additional information about the watermark payload $b$. Third, the editor performs no adaptation informed by the decoder. With payloads sampled independently of $C$, the second condition implies that $I(b;Z)$ is small. These assumptions separate two quantities that robustness evaluations often conflate: a model can preserve information about depicted content while discarding information about the forensic payload embedded in the image~\citep{moulin2003information,vanderoord2017vqvae,rombach2022ldm}.

\begin{proposition}[Reconstruction bottleneck]
\label{prop:bottleneck}
For the Markov pathway $b\rightarrow x_V\rightarrow Z\rightarrow y$, data processing gives
\begin{equation}
I(b;y\mid p)\leq I(b;Z).
\label{eq:bottleneck}
\end{equation}
Thus, a representation that retains little information about the victim payload cannot support reliable downstream recovery, even when it preserves visible content. The next proposition states the corresponding BER implication.
\end{proposition}

\begin{proposition}[Implication for optimal recovery]
\label{prop:fano}
If the payload bits $b_1,\ldots,b_k$ are independent and uniform, let $P_{e,j}^*$ be the minimum error probability over all estimators of $b_j$ from $y$. Then
\begin{equation}
\BER^*(b\mid y):=\frac{1}{k}\sum_{j=1}^{k}P_{e,j}^*\geq
\frac{1}{k}\sum_{j=1}^{k}h_2^{-1}\!\left(1-I(b_j;y)\right).
\label{eq:fano}
\end{equation}
Here, $h_2(q)=-q\log_2 q-(1-q)\log_2(1-q)$ is the binary entropy function, and $h_2^{-1}$ denotes its inverse restricted to $q\in[0,0.5]$. For a uniform payload bit, $1-I(b_j;y)$ represents the residual uncertainty about $b_j$ after observing $y$. Together with $P_{e,j}^*\leq0.5$, the bound implies that vanishing information about every bit forces optimal recovery toward the limit of random decoding. This statement concerns recoverable information. The empirical tables instead report the raw BER of each fixed victim decoder and interpret it by proximity to $0.5$. The result follows from the bitwise form of Fano's inequality and the local expansion of binary entropy~\citep{cover2006elements}.
\end{proposition}

The bottleneck also predicts invariance to prompts. If prompts affect only the renderer and do not introduce a new information channel from $b$ beyond $Z$, then changing prompt wording can alter color, texture, sharpness, or geometry but cannot recover payload information already discarded by the representation. Explicit ``remove hidden information'' language is therefore not necessary for BER to approach $0.5$. This claim is conditional because prompts can still determine whether the result remains faithful enough to qualify as laundering.

For residual-sensitive schemes, we measure normalized high-frequency change
\begin{equation}
\HFR(x_V,y)=
\frac{\lVert P_H(y)-P_H(x_V)\rVert_2}
{\lVert P_H(x_V)\rVert_2+\eta},
\label{eq:hfr}
\end{equation}
where $P_H$ selects high DCT bands. Large changes in the high-frequency residual can cause decoder margins to cross their decision boundaries. HFR summarizes the magnitude of these changes, but it need not determine BER because schemes differ in payload placement, redundancy, and decoder invariances.

The evaluation tests five linked expectations. Payload disruption can coexist with fidelity (H1), and schemes can respond asymmetrically because their carriers and decoders differ (H2). Prompt components should affect fidelity more than BER when payload loss occurs before rendering (H3). Later model versions can improve fidelity while preserving greater payload recoverability (H4). HFR should be positively associated with raw BER within the primary grid (H5). Because BER is not monotonic as an attack metric beyond $0.5$, this association is interpreted as a relationship with decoder error rather than a monotonic disruption score. The theoretical account would be weakened if movement toward random recovery required explicit removal wording, occurred only when semantic fidelity collapsed, or showed no residual association for any evaluated scheme.

\section{Experimental Setup}
\label{sec:setup}

\paragraph{Models and watermarks.}
We evaluate GPT Image~1, GPT Image~1.5, and GPT Image~2, together with Nano Banana, Nano Banana Pro, and Nano Banana~2~\citep{gptimage1,gptimage2,nanobananapro,nanobanana2}. The three victim watermarking schemes are DwtDct~\citep{barni1998dct}, DwtDctSvd~\citep{kumari2023dwtsvd}, and RivaGAN~\citep{zhang2019rivagan}. All clean references are resized to $1024\times1024$ before watermark embedding, and all calls use the public image editing interfaces at that resolution. Before every API call, we retain only inputs whose payload is correctly decoded, preventing embedding failures from being counted as laundering.

\paragraph{Data and protocol.}
The stratified target list contains 100 MS-COCO images~\citep{lin2014coco}: 40 scenes rich in texture, 30 scenes with large flat regions, and 30 scenes containing text. The full factorial grid has $3\times6=18$ watermark--model cells and 100 outputs per cell, for 1,800 primary API calls. A transient failure before output generation is retried once on the same image. A persistent failure triggers replacement with the next unused image from the same stratum, independently of decoder performance. Thus, every cell preserves the same stratum counts, although a rare replacement can break pairing of individual images across editing models. Every cell uses the same structured reconstruction prompt $P_{\rm struct}$ without prompt search, repeated editing, or detector feedback. The full structured prompt and minimal prompt are reproduced in the supplementary material.

\begin{figure*}[t]
\centering
\includegraphics[width=.98\textwidth]{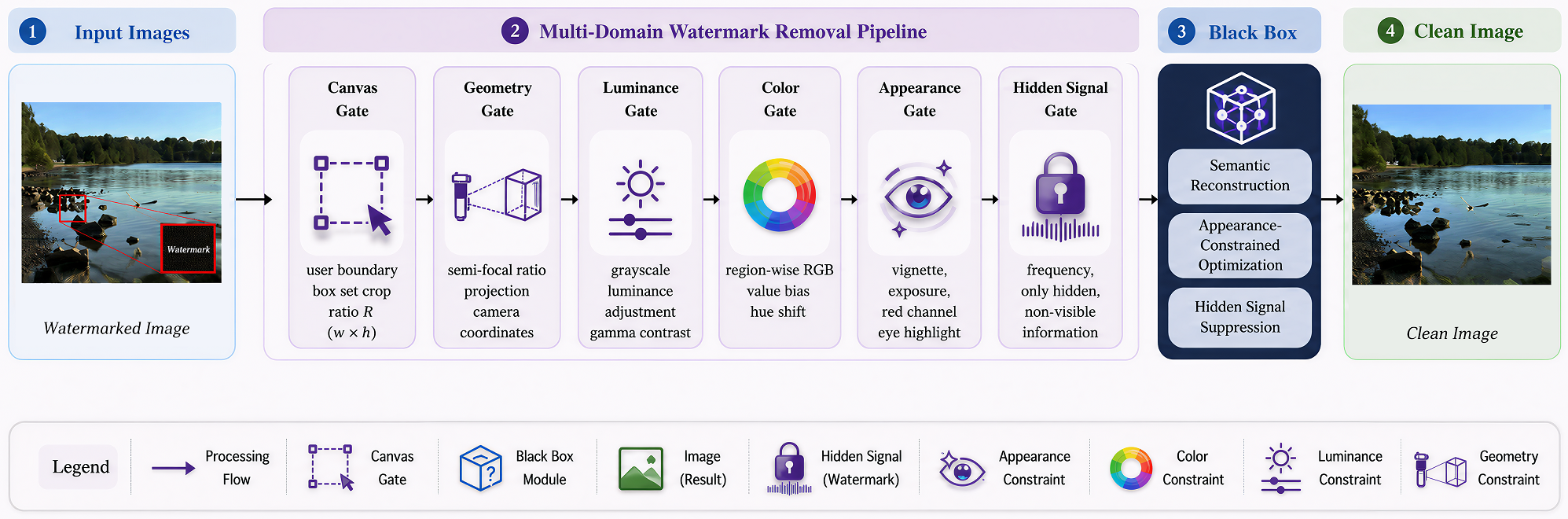}
\caption{Structure of the single prompt $P_{\rm struct}$. The six gates represent prompt clauses for canvas, geometry, luminance, color, appearance, and hidden-signal handling before a black-box reconstruction call; they describe the request rather than observed internal model stages.}
\label{fig:prompt-structure}
\end{figure*}

The structured prompt asks the editor to preserve objects, layout, canvas size, luminance, color statistics, and regional geometry while returning a single reconstructed image. It also includes preservation of visible content, grayscale luminance guidance, regional color constraints, removal of hidden information, and final output validation. These instructions define a constrained reconstruction request rather than an optimization loop: there is one input, one prompt, and one returned output. Failure handling never uses the performance of the victim decoder. Figure~\ref{fig:prompt-structure} visualizes the modular organization of $P_{\rm struct}$.

\paragraph{Ablations, metrics, and baselines.}
The main evaluation covers DwtDct, DwtDctSvd, and RivaGAN, whereas the focused prompt ablation is conducted on DwtDct only, using Nano Banana~2 and GPT Image~2. The tested variants separately remove preservation of visible content, joint luminance and color guidance, appearance and geometry constraints, and explicit language about hidden information. The evaluation also includes a preservation constraint and a minimal prompt. We report raw victim BER, for which proximity to $0.5$ indicates stronger disruption. Reference fidelity is measured with PSNR, SSIM~\citep{wang2004ssim}, LPIPS~\citep{zhang2018lpips}, and CLIP-Sim~\citep{radford2021clip}. Table~\ref{tab:baseline-compact} reports clean-reference and no-reference image quality metrics. Conventional comparisons include Gaussian blur, Gaussian noise, contrast scaling, $30^\circ$ rotation, 50\% cropping, and BM3D~\citep{dabov2007bm3d}. Means are reported over the stratified evaluation design. Formal uncertainty testing remains a limitation.

Reference and semantic metrics assess whether the output remains tied to the submitted image. No-reference metrics instead assess image quality without direct comparison to the source. We keep these metric families separate because a model may ``repair'' blur, darkness, or compression artifacts and obtain a better no-reference score while moving farther from the source under PSNR or SSIM. Likewise, BER is never interpreted alone: a value near the limit of random recovery does not provide strong laundering evidence when the output is unrecognizable under Definition~\ref{def:watermark-laundering}.

\section{Results}
\label{sec:results}

\subsection{Joint BER--Fidelity Profile}

Table~\ref{tab:core-results} reports the main experimental results for all 18 watermark--model configurations under $P_{\rm struct}$, with $N=100$ outputs per configuration. It jointly presents victim-payload disruption and reconstruction fidelity rather than reducing the two objectives to a single score. The GPT models occupy the regime with stronger disruption and lower reference fidelity. GPT Image~1 has the mean BER closest to random recovery (0.4808). GPT Image~2 improves mean PSNR over the earlier GPT models while retaining substantial disruption, with BER values of 0.4266, 0.4517, and 0.3620 for DwtDct, DwtDctSvd, and RivaGAN, respectively. The Google models yield progressively higher fidelity. Nano Banana~2 reaches a mean PSNR of 30.27 and a mean SSIM of 0.879 against the watermarked input, while DwtDct remains substantially disrupted at BER 0.4110. This high-fidelity laundering condition supports H1 within the evaluated setting because payload disruption persists without obvious image destruction.

\begin{table}[t]
\centering
\caption{Core laundering results under $P_{\rm struct}$. Panel (a) reports BER for each victim watermark; values closer to $0.5$ indicate stronger payload disruption. Panel (b) reports fidelity metrics averaged across the three watermarks, using the watermarked input as the reference. Each watermark--model cell has $N=100$.}
\label{tab:core-results}
\small
\setlength{\tabcolsep}{3pt}
\begin{tabular}{lrrr}
\multicolumn{4}{c}{\textit{(a) Payload disruption (BER$\rightarrow0.5$)}} \\
\toprule
Editing model & DwtDct & DwtDctSvd & RivaGAN \\
\midrule
GPT Image~1   & \textbf{0.4944} & 0.4907 & 0.4574 \\
GPT Image~1.5 & 0.4353 & \textbf{0.4963} & \textbf{0.4649} \\
GPT Image~2   & 0.4266 & 0.4517 & 0.3620 \\
Nano Banana   & 0.4106 & 0.4167 & 0.1572 \\
Nano Banana Pro & 0.4159 & 0.2947 & 0.1717 \\
Nano Banana~2 & 0.4110 & 0.2658 & 0.1233 \\
\bottomrule
\end{tabular}

\vspace{3pt}

\begin{tabular}{lrrr}
\multicolumn{4}{c}{\textit{(b) Reconstruction fidelity}} \\
\toprule
Editing model & PSNR$\uparrow$ & SSIM$\uparrow$ & CLIP-Sim$\uparrow$ \\
\midrule
GPT Image~1   & 14.0360 & 0.4051 & 0.8939 \\
GPT Image~1.5 & 15.0732 & 0.4069 & 0.9056 \\
GPT Image~2   & 16.9877 & 0.5017 & 0.9424 \\
Nano Banana   & 20.4589 & 0.5738 & 0.9682 \\
Nano Banana Pro & 25.8871 & 0.7497 & 0.9429 \\
Nano Banana~2 & \textbf{30.2726} & \textbf{0.8794} & \textbf{0.9837} \\
\bottomrule
\end{tabular}
\end{table}

The table does not support an explanation that is independent of the watermark scheme. DwtDct remains at a BER of approximately 0.41 across all six editing models, including the high-fidelity Google models. DwtDctSvd is similarly vulnerable to the earlier models but becomes more recoverable under Nano Banana Pro and Nano Banana~2. RivaGAN shows the largest model dependence: its learned decoder is strongly disrupted by the GPT models but retains substantially more payload under the later Google models. These differences are consistent with H2. The pattern across schemes is inconsistent with a single explanation based on frequency bands and is compatible with differences in payload placement, redundancy, and decoder invariances.

For both providers, later model versions improve source fidelity while mean BER moves farther from $0.5$. Mean BER changes from 0.4808 to 0.4655 to 0.4134 across GPT Image versions and from 0.3282 to 0.2941 to 0.2667 across Nano Banana versions. This trend supports H4: later versions can improve fidelity while preserving greater payload recoverability rather than making laundering monotonically stronger. DwtDct is the exception, remaining at a BER of approximately 0.41 across all Google models despite the large fidelity gain.

\begin{figure}[t]
\centering
\includegraphics[width=\columnwidth]{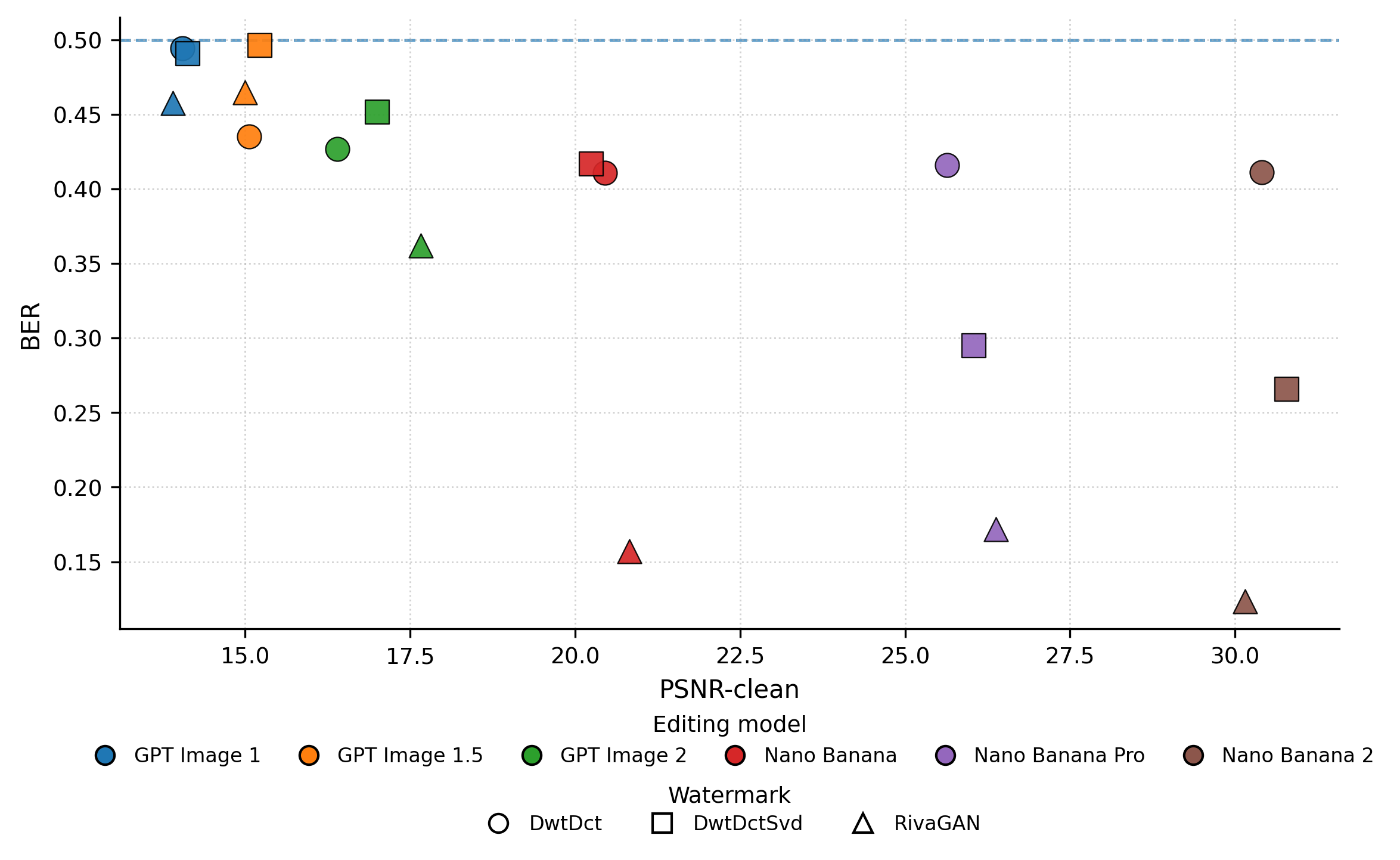}
\caption{Victim BER versus clean-reference PSNR. Marker shape denotes the watermark scheme. The dashed line marks random binary recovery.}
\label{fig:ber-fidelity-scatter}
\end{figure}

Figure~\ref{fig:ber-fidelity-scatter} presents the two objectives using clean-reference PSNR. Points near the dashed BER $=0.5$ line indicate payload recovery approaching the random limit, while movement to the right indicates greater source fidelity. GPT models cluster in a region of high disruption and low fidelity. Nano Banana~2 moves substantially to the right, although its DwtDct point remains above BER 0.4. A ranking based only on BER would therefore obscure this high-fidelity condition.

Figure~\ref{fig:attack-grid}(a) illustrates the two regimes. GPT Image~2 preserves scene semantics but visibly changes exposure, contrast, texture, boundaries, and regions that resemble text, whereas Nano Banana~2 remains closely aligned with the source. The no-reference image quality metrics in Table~\ref{tab:baseline-compact} sometimes favor the more extensively altered outputs. This result reflects a mismatch between perceived naturalness and source fidelity rather than contradicting the lower fidelity at the pixel level.

\subsection{Prompt Sensitivity of Watermark Laundering}

\begin{figure*}[!t]
\centering
\begin{minipage}[t]{.58\textwidth}
\centering
\includegraphics[width=\linewidth]{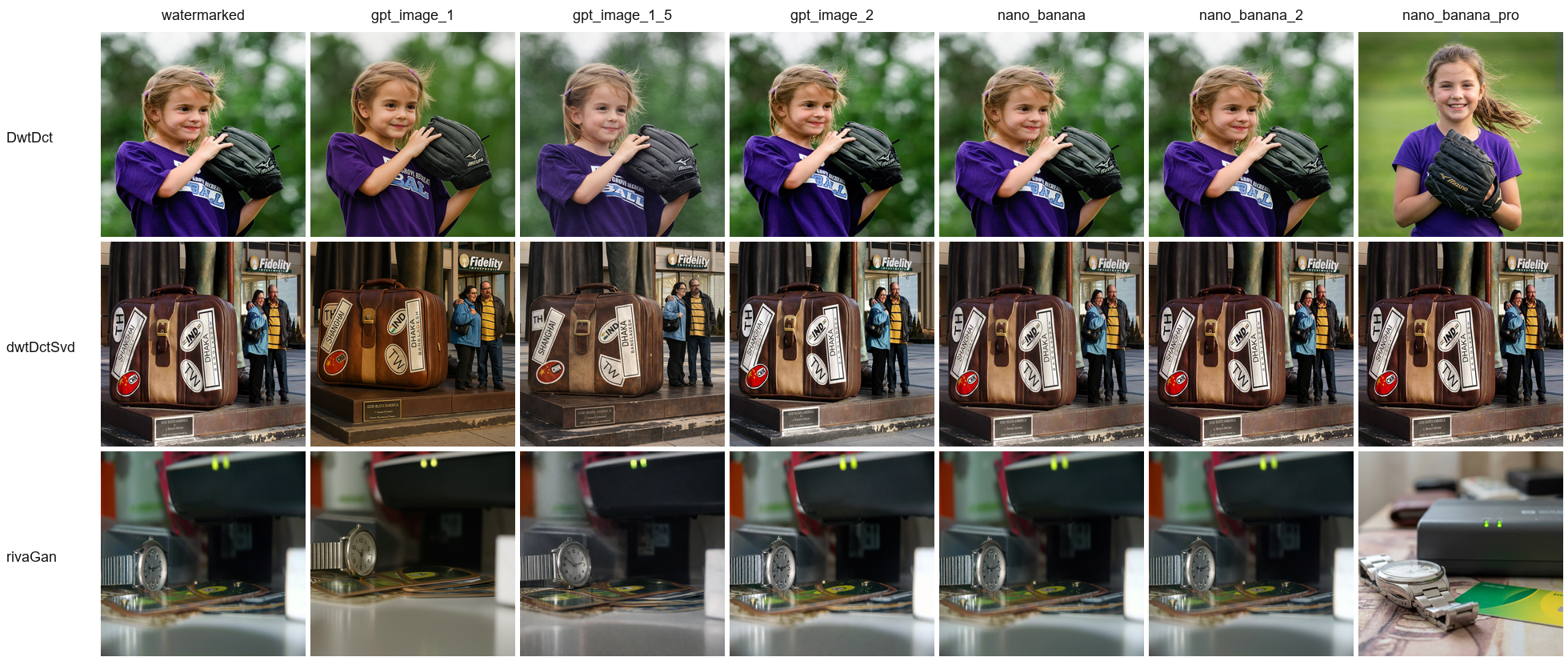}
\end{minipage}\hfill
\begin{minipage}[t]{.38\textwidth}
\centering
\includegraphics[width=\linewidth]{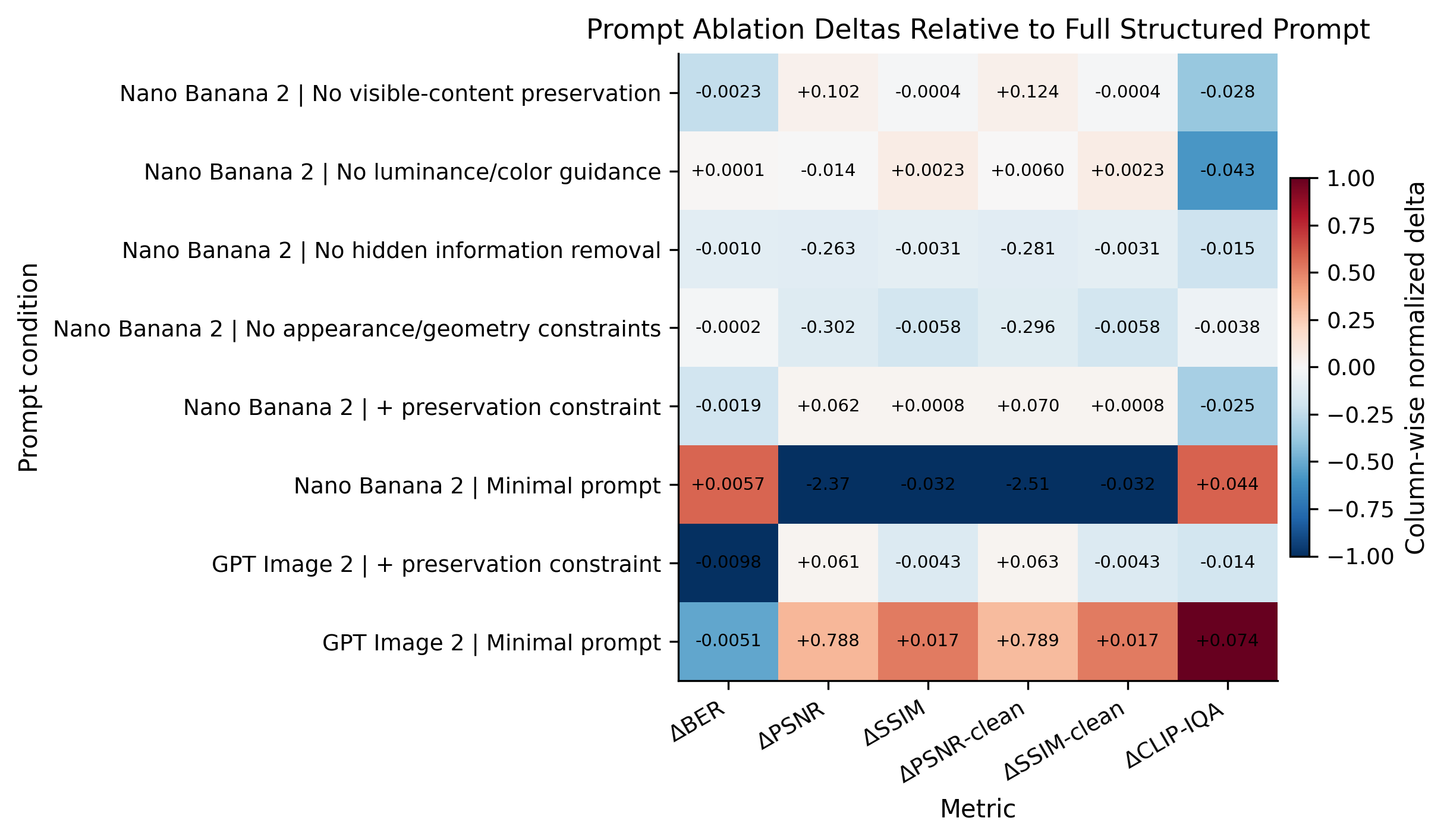}
\end{minipage}
\caption{(a) Representative outputs generated with one prompt across editing models and victim watermark schemes. (b) Prompt-ablation changes relative to $P_{\rm struct}$ for DwtDct; positive $\Delta$BER moves toward random recovery because all evaluated BER values are below $0.5$.}
\label{fig:attack-grid}
\label{fig:prompt-ablation-heatmap}
\end{figure*}

Figure~\ref{fig:prompt-ablation-heatmap}(b) summarizes the DwtDct ablation. On Nano Banana~2, removing any tested module keeps BER within 0.0023 of the value of 0.4110 under the full prompt. This result is consistent with payload disruption not requiring explicit language about hidden information. Fidelity is more sensitive. Removing the appearance and geometry constraints lowers PSNR by 0.302. Removing the joint luminance and color guidance increases SSIM by 0.0023 and decreases LPIPS, with essentially unchanged BER. The minimal prompt lowers Nano Banana~2 PSNR by 2.369. For GPT Image~2, it increases PSNR by 0.788 while changing BER by only $-0.0051$. The contrast between stable BER and larger fidelity changes supports H3. These descriptive differences are not treated as statistically resolved effects because formal uncertainty testing is absent.

The luminance and color guidance ablation is descriptive rather than a direct observation of model internals. On Nano Banana~2, omitting the grayscale luminance anchor and region-wise color constraint improves SSIM by 0.0023 and produces the lowest LPIPS among its ablations without changing BER. Because the editors are closed source, this result describes how the requested constraints affect the returned reconstruction. It does not establish that the models instantiate the requested intermediate representations. The stable BER is consistent with the editing pathway contributing to payload disruption without any single prompt module being necessary.

The comparison with the minimal prompt also constrains the policy interpretation. Explicit removal language is not required to disrupt the victim payload. The minimal request reaches BER 0.4167 for Nano Banana~2 and 0.4215 for GPT Image~2. The relevant security concern therefore extends beyond a single prohibited phrase that a lexical filter could block. General reconstruction capability can produce the transformation even when the request is phrased as ordinary image recreation. This result does not demonstrate a bypass of provider safeguards, but it indicates that lexical filtering alone would not address the observed reconstruction behavior.

\subsection{Comparison with Conventional Attacks}

Table~\ref{tab:baseline-compact} combines payload disruption, clean-reference fidelity, and no-reference image quality for the retained conventional baselines and two laundering conditions produced by foundation models. For DwtDct, contrast scaling gives the BER closest to $0.5$. The BER under Gaussian noise is 0.6225, which lies 0.1225 above the point of random recovery and coincides with low clean-reference fidelity and weaker no-reference quality. GPT Image~2 gives the BER closest to $0.5$ for DwtDctSvd and RivaGAN. It also gives the best NIQE, BRISQUE, and CLIP-IQA values within every displayed watermark block, which is consistent with automatic repair behavior. Nano Banana~2 provides the laundering regime with higher fidelity, whereas Gaussian blur and BM3D preserve the clean reference most closely but are less consistently disruptive. The comparison does not exhaust regeneration methods. Its purpose is to distinguish the public reconstruction interface from fixed local operators.

\begin{table*}[!t]
\centering
\caption{BER and image-quality comparison. Background colors distinguish victim watermark schemes. PSNR-clean and SSIM-clean use the clean reference; NIQE, BRISQUE, and CLIP-IQA are no-reference metrics. Bold marks the BER closest to $0.5$ or the best quality value within each color block.}
\label{tab:baseline-compact}
\footnotesize
\setlength{\tabcolsep}{3.5pt}
\renewcommand{\arraystretch}{0.84}
\begin{tabular}{lrrrrrr}
\toprule
Attack condition & BER$\rightarrow0.5$ & PSNR-clean$\uparrow$ & SSIM-clean$\uparrow$ & NIQE$\downarrow$ & BRISQUE$\downarrow$ & CLIP-IQA$\uparrow$ \\
\midrule
\rowcolor{DwtDctShade}\multicolumn{7}{l}{\textit{DwtDct}} \\
\rowcolor{DwtDctShade}Gaussian blur & 0.3134 & 35.6690 & \textbf{0.9631} & 1.6203 & 50.4784 & 0.6505 \\
\rowcolor{DwtDctShade}Gaussian noise & 0.6225 & 20.4483 & 0.4112 & 1.9251 & 58.6475 & 0.4967 \\
\rowcolor{DwtDctShade}Contrast & \textbf{0.4587} & 14.2076 & 0.6465 & 2.2275 & 58.5809 & 0.4370 \\
\rowcolor{DwtDctShade}Rotation & 0.4094 & 9.7087 & 0.3037 & 1.4469 & 28.3639 & 0.2276 \\
\rowcolor{DwtDctShade}Cropping & 0.0356 & 10.0902 & 0.2796 & 1.4783 & 45.3134 & 0.4150 \\
\rowcolor{DwtDctShade}BM3D denoise & 0.4097 & \textbf{36.7900} & 0.9423 & 1.6544 & 64.1586 & 0.6470 \\
\rowcolor{DwtDctShade}GPT Image~2 & 0.4266 & 16.3935 & 0.4758 & \textbf{1.4092} & \textbf{18.4072} & \textbf{0.7361} \\
\rowcolor{DwtDctShade}Nano Banana~2 & 0.4110 & 30.4004 & 0.8769 & 1.4113 & 20.1624 & 0.5394 \\
\addlinespace
\rowcolor{DwtDctSvdShade}\multicolumn{7}{l}{\textit{DwtDctSvd}} \\
\rowcolor{DwtDctSvdShade}Gaussian blur & 0.0000 & 35.1716 & \textbf{0.9632} & 1.6243 & 50.2400 & 0.5898 \\
\rowcolor{DwtDctSvdShade}Gaussian noise & 0.1331 & 20.3887 & 0.3921 & 1.9254 & 58.9234 & 0.4717 \\
\rowcolor{DwtDctSvdShade}Contrast & 0.3787 & 14.3713 & 0.6686 & 2.2672 & 58.6463 & 0.3656 \\
\rowcolor{DwtDctSvdShade}Rotation & 0.4062 & 9.8574 & 0.3104 & 1.4467 & 23.5672 & 0.1996 \\
\rowcolor{DwtDctSvdShade}Cropping & 0.0000 & 10.2587 & 0.2909 & 1.4771 & 44.1477 & 0.3526 \\
\rowcolor{DwtDctSvdShade}BM3D denoise & 0.3591 & \textbf{36.2090} & 0.9412 & 1.6607 & 63.8982 & 0.6128 \\
\rowcolor{DwtDctSvdShade}GPT Image~2 & \textbf{0.4517} & 16.9978 & 0.5135 & \textbf{1.3907} & \textbf{16.3967} & \textbf{0.7036} \\
\rowcolor{DwtDctSvdShade}Nano Banana~2 & 0.2658 & 30.7813 & 0.8847 & 1.3948 & 19.8320 & 0.4452 \\
\addlinespace
\rowcolor{RivaGANShade}\multicolumn{7}{l}{\textit{RivaGAN}} \\
\rowcolor{RivaGANShade}Gaussian blur & 0.0000 & 35.2932 & \textbf{0.9631} & 1.6125 & 50.7863 & 0.5548 \\
\rowcolor{RivaGANShade}Gaussian noise & 0.0591 & 20.4930 & 0.4015 & 1.9186 & 58.4579 & 0.4425 \\
\rowcolor{RivaGANShade}Contrast & 0.1309 & 14.3946 & 0.6524 & 2.2565 & 57.0378 & 0.3511 \\
\rowcolor{RivaGANShade}Rotation & 0.3016 & 9.9658 & 0.3011 & 1.4265 & 26.4614 & 0.2225 \\
\rowcolor{RivaGANShade}Cropping & 0.0013 & 10.0768 & 0.2683 & 1.4234 & 44.4159 & 0.3787 \\
\rowcolor{RivaGANShade}BM3D denoise & 0.0191 & \textbf{36.2624} & 0.9415 & 1.6474 & 62.1148 & 0.6002 \\
\rowcolor{RivaGANShade}GPT Image~2 & \textbf{0.3620} & 17.6600 & 0.5171 & \textbf{1.3672} & \textbf{17.2455} & \textbf{0.6726} \\
\rowcolor{RivaGANShade}Nano Banana~2 & 0.1233 & 30.1518 & 0.8774 & 1.3760 & 18.5228 & 0.5033 \\
\bottomrule
\end{tabular}
\end{table*}

\subsection{High-Frequency Residual Analysis}

Across 1,800 matched input--output pairs, HFR is positively associated with raw BER (Pearson $r=0.3158$, Spearman $\rho=0.3607$). This statistic describes decoder error and is not a monotonic measure of disruption for individual observations with BER above $0.5$. The association is positive within DwtDct ($r=0.3361$), DwtDctSvd ($r=0.3014$), and RivaGAN ($r=0.4662$), but it is weak within several model aggregates. The aggregate association supports H5, while the weak within-model results limit HFR to one contributor to decoder errors rather than a complete scalar explanation. The measured change in high frequencies does not identify the coefficients that carry the payload or the learned invariances of each decoder.

The means for each model indicate a change in reconstruction behavior. Mean HFR decreases from 7.3947 for GPT Image~1 to 3.4389 for GPT Image~1.5 and 2.5204 for GPT Image~2, while mean BER changes from 0.4808 to 0.4655 to 0.4134. Later OpenAI models therefore produce smaller mean changes in the high-frequency residual while also moving farther from random bit recovery. Nano Banana Pro has the strongest association between HFR and BER within a model ($r=0.3585$), whereas the correlations for the GPT models are near zero. HFR measures the magnitude of residual change but not whether that change aligns with the effective decision directions of a particular decoder.

GPT Image~2 also exhibits apparent repair behavior. Dark, blurry, compressed, or low-contrast inputs may be brightened, sharpened, or locally reconstructed. Such outputs can appear more natural and score better on no-reference quality metrics even when PSNR and SSIM decrease against the source. No-reference metrics assess perceived naturalness, whereas PSNR and SSIM assess source fidelity. The high-fidelity Nano Banana~2 results further indicate that laundering is not confined to visibly repaired or aggressively reconstructed images.

\section{Security Implications for Watermark Laundering}
\label{sec:security-implications}

Watermark laundering indicates a mismatch between model safety checks and provenance security. An editor may reject explicit requests to remove a watermark or alter ownership evidence while accepting a visually benign instruction to recreate, restore, or clean an image. If the reconstruction pathway discards the victim carrier, compliance with a policy defined at the level of wording does not guarantee provenance preservation in the output. An attacker can therefore use the prompt to repurpose a permitted editing operation for an effect that is relevant to provenance. The ablation results indicate that phrase blocking alone is insufficient because explicit language about removing hidden information is not necessary for the observed disruption.

The stronger state in Equation~\eqref{eq:provenance-substitution} further clarifies the risk. Our provider marker audit found C2PA/Content Credentials binary markers in all 100 sampled outputs from each provider~\citep{c2pa_spec,openai_c2pa_images,gemini_imagegen}. A compatible signature verifier was unavailable, and SynthID was not tested. The audit therefore indicates a provider-side marker pathway but does not establish $E_{\rm prov}=1$ or contribute to the laundering success criterion. If recovery of the victim payload becomes unreliable and the editor attaches verified provenance evidence, the returned file can present an apparent machine-readable origin that differs from the provenance of the victim image. An attacker could also combine reconstruction with ordinary metadata or changes to visible labels to create an apparent copyright or attribution replacement. These operations constitute attacks on provenance and ownership evidence, but they do not change the underlying legal rights. Defenses should therefore bind authorization and provenance across transformations, preserve or cryptographically link upstream claims, and test outputs for loss of the victim signal rather than relying only on prompt moderation or the presence of a new provider mark.

\section{Discussion and Conclusion}
\label{sec:conclusion}

Within the evaluated setting, the results demonstrate that watermark laundering is a practical failure mode for victim watermarks: a single prompt submitted to a public black-box editor can disrupt invisible payloads while returning usable reconstructions. The OpenAI models provide the strongest disruption across schemes. GPT Image~1 is closest to random recovery on average, while GPT Image~2 improves fidelity and retains substantial disruption. The Nano Banana~2 results indicate that vulnerability in the transform domain can persist at substantially higher fidelity. The prompt ablations, conventional baselines, version comparisons, and HFR measurements are consistent with the reconstruction-bottleneck account, but they do not verify the internal causal pathways of proprietary editors. Watermark robustness is therefore not equivalent to provenance security. Future evaluations should include reconstruction with foundation models and prompts alongside classical perturbation, denoising, and generative reconstruction baselines. They should also report raw BER, interpret it by proximity to $0.5$, and pair it with reference and semantic fidelity.

The evidence is limited to three watermark schemes, six evaluated editing models, a stratified design with 100 images, one prompt family, and a conventional baseline suite that does not include an additional diffusion reconstruction method. Formal confidence intervals and tests of effect size are also absent. We do not claim coverage of generative watermarks embedded within models, such as Stable Signature, Tree-Ring, or Gaussian Shading~\citep{fernandez2023stablesig,wen2023treering,yang2024gaussianshading}. We also do not claim a bypass of provider policy, legal copyright transfer, or universal provenance substitution. Because model versions and interfaces evolve, the reported values constitute a temporal snapshot. The conclusion is limited to the evaluated interface: reconstruction accessed through prompts can decouple preservation of visible content from recovery of the victim payload. Watermark defenses and model safety designs should therefore evaluate that interface directly.

\section*{Acknowledgments}
This work was supported in part by Taishan Scholar under Grant tsqnz20250747; in part by the National Natural Science Foundation under Grant 62502250, Grant 62406051, Grant 62302249, Grant 62541206, and Grant 62272255; and in part by the Young Talent of Lifting Engineering for Science and Technology in Shandong under Grant SDAST2025QTB030.

\begingroup
\small
\bibliographystyle{unsrt}
\bibliography{paper_references}
\endgroup

\end{document}